%% file: main.tex
\PassOptionsToPackage{table,xcdraw}{xcolor}
\documentclass[sigconf]{acmart} 

\usepackage{xcolor} 
\usepackage{colortbl} 
\AtBeginDocument{%
  }
\setcopyright{acmlicensed}
\copyrightyear{2025}
\acmYear{2025}
\acmDOI{XXXXXXX.XXXXXXX}
\acmConference[Preprint, Accepted at SLTAT'25]{ACM International Conference on Intelligent Virtual Agents (IVA'25)}{Sept 16--19,2025}{Berlin, Germany}
\acmISBN{978-1-4503-XXXX-X/2018/06}

\begin{document}

\title{Contrastive Pretraining with Dual Visual Encoders for Gloss-Free Sign Language Translation}


\author{Ozge Mercanoglu Sincan}
\email{o.mercanoglusincan@surrey.ac.uk}
\affiliation{%
  \institution{CVSSP, University of Surrey}
  \city{Guildford}
  \country{UK}
}

\author{Richard Bowden}
\email{r.bowden@surrey.ac.uk}
\affiliation{%
  \institution{CVSSP, University of Surrey}
  \city{Guildford}
  \country{UK}
}

\renewcommand{\shortauthors}{Sincan et al.}


\begin{abstract}
Sign Language Translation (SLT) aims to convert sign language videos into spoken or written text. While early systems relied on gloss annotations as an intermediate supervision, such annotations are costly to obtain and often fail to capture the full complexity of continuous signing. In this work, we propose a two-phase, dual visual encoder framework for gloss-free SLT, leveraging contrastive visual–language pretraining. During pretraining, our approach employs two complementary visual backbones whose outputs are jointly aligned with each other and with sentence-level text embeddings via a contrastive objective. During the downstream SLT task, we fuse the visual features and input them into an encoder–decoder model. On the Phoenix-2014T benchmark, our dual encoder architecture consistently outperforms its single-stream variants and achieves the highest BLEU-4 score among existing gloss-free SLT approaches.
\end{abstract}

\ccsdesc[500]{Human-centered computing~Accessibility technologies}

\keywords{Sign Language Translation, Gloss-free SLT, Contrastive Learning, Multimodal Pretraining}


\maketitle
   
\input{paper_sec/1_intro}

\input{paper_sec/2_related}

\input{paper_sec/3_method}

\input{paper_sec/4_experiments}
\input{paper_sec/5_conclusion}

\bibliographystyle{ACM-Reference-Format}
\bibliography{main}

\end{document}

%% file: paper_sec/1_intro.tex
\section{Introduction}
Sign languages are visual languages that rely on both manual (handshape, orientation, movement) and non-manual (facial expressions, head and body posture) components. These components operate simultaneously and often encode grammatical information in parallel channels. With over 200 sign languages in the world \cite{wfdeaf}, the development of inclusive technologies to support sign language understanding has become an increasingly important research domain.

Sign Language Translation (SLT) aims to convert sign language videos into spoken or written text. Achieving this requires the models to capture both fine-grained spatial detail (e.g., specific handshapes and finger configurations), temporal dynamics (e.g., motion trajectories). These visual features are then transformed into coherent natural language output. Works in this domain have predominantly relied on sequential models, such as Recurrent Neural Networks (RNNs) or the Transformer \cite{vaswani2017attention}, to map sign language video sequences to spoken language sentences \cite{camgoz2018neural, camgoz2020sign, sincan2023context, yao2023sign, zhang2022sltunet}.  Earlier studies relied on intermediate annotations called glosses -- simplified written descriptions of individual signs -- to act as a bridge between visual and linguistic domains \cite{camgoz2018neural, camgoz2020sign, zhang2022sltunet}. However, manual annotation of glosses is labor-intensive and difficult to annotate accurately, resulting in a limited number of publicly available gloss-annotated datasets \cite{dgscorpus_3}.

To overcome this limitation, gloss-free SLT methods have been proposed, aiming to directly translate sign language videos into spoken language without relying on gloss supervision \cite{chen2024factorized, gong2024llms, ye2024improving, zhou2023gloss}. This shift removes the dependence on expert gloss annotations but brings challenges in learning rich, discriminative representations of signing sequences. Typical architectures use a frozen visual encoder backbone, such as 3D Convolutional Neural Networks, e.g., I3D \cite{carreira2017quo}, to extract features, and pass them into transformers \cite{albanie2021bbc, shi2022ttic, sincan2023context}. However, these models often lack a strong semantic bridge between visual and linguistic modalities, limiting their ability to generate semantically accurate translations.

Recent advancements in gloss-free SLT have drawn inspiration from visual-language pretraining methods \cite{zhou2023gloss, ye2024improving, chen2024factorized, wong2024signgpt}. Approaches such as GFSLT-VLP \cite{zhou2023gloss} pioneered the use of contrastive learning to align visual representations with textual descriptions. These methods offer a promising direction to improve the semantic grounding of visual features and provide more flexible and scalable training paradigms for SLT.

In this work, we propose a dual visual encoder framework for gloss-free SLT. We employ two complementary visual backbones: a ResNet \cite{he2016deep} to extract fine-grained spatial features from individual frames, and an I3D \cite{carreira2017quo} to capture spatio-temporal features. During our pretraining stage, both feature sets are projected into a shared embedding space and aligned via a contrastive loss against corresponding sentence embeddings extracted by a pretrained language model. We additionally include an inter-modal alignment that encourages the ResNet and I3D features from the same video to be similar while pushing apart features from different video samples. After pretraining, we fine-tune a translation model for translation by fusing visual features and feeding them into an mBART-based encoder-decoder architecture. We evaluate our approach on the Phoenix-2014T dataset \cite{camgoz2018neural} and demonstrate significant improvements over our single-encoder baselines. We outperform existing gloss-free SLT methods on the BLEU-4 metric, while obtaining competitive results on other evaluation metrics.
    \begin{itemize}
        \item We introduce a contrastively pretrained dual visual encoder architecture that jointly aligns both visual modalities with each other and with language. This alignment encourages the model to learn complementary representations and enhance semantic consistency.
        \item We explore and compare various fusion strategies for combining visual features.
    \end{itemize}

%% file: paper_sec/2_related.tex
\section{Related Work}

Recent research at the intersection of vision and language has led to rapid advances in many tasks, including sign language translation. In this section, we review works on visual-language pretraining and sign language translation.

\subsection{Visual-Language Pretraining}

Contrastive pretraining between vision and language has become fundamental for many downstream tasks. A pioneering work in this field, CLIP (Contrastive Language-Image Pre-training) \cite{radford2021learning} aligns image features with text representations using a contrastive learning strategy. CLIP does not need task-specific training data, and it demonstrates an ability to generalize across a wide range of downstream tasks such as zero-shot transfer to image classification, facial emotion recognition, optical character recognition, and so on. VideoCLIP \cite{xu2021videoclip} extends this contrastive learning objective to pre-train a unified video-text representation that surpasses prior works in video understanding tasks such as text-video retrieval, video question answering, and action segmentation. 

Several extensions have adapted this idea for additional modalities. CLIP4VLA \cite{ruan2023accommodating} extends CLIP by incorporating audio as a third modality, forming a unified tri-encoder structure for multimodal learning. IMAGEBIND \cite{girdhar2023imagebind} scales this idea further by binding six modalities (e.g., image, text, video, audio, depth, and thermal) -- using images as a key modality, and aligning each modality to image embeddings. More recently, LANGUAGEBIND \cite{zhu2024languagebind} proposes a language-based multi-modal pretraining, taking the language as the bind across video, infrared, depth, and audio.

\subsection{Sign Language Translation}

Sign language translation presents unique challenges compared to other vision-language tasks due to its complexity. Early studies \cite{camgoz2018neural, camgoz2020sign, chen2022two, zhang2022sltunet, chen2022simple} decomposed the SLT task two sub-tasks:  Sign Language Recognition (SLR), recognizing sequences of signs, and then `translating' the recognized signs into spoken language sentences. These gloss-based pipelines benefit from structured intermediate supervision, enabling models to focus on smaller vocabularies and more discrete targets. However, gloss annotations are highly labor-intensive and not available for large datasets, motivating gloss-free SLT methods. 

Gloss-free SLT approaches aim to directly translate sign language videos into spoken language without gloss annotation. Initial gloss-free efforts (Sign2Text) typically adopted end-to-end encoder–decoder frameworks using a visual encoder and transformer-based decoders \cite{camgoz2020sign}. While these architectures removed the need for glosses, their performance was often inferior to gloss-based counterparts, largely due to weaker semantic grounding between visual features and linguistic output.

To address this, several works have focused on enhancing visual representations. Some approaches \cite{shi2022open, sincan2023context} pretrain a classifier on sign language datasets and use it as a frozen visual encoder for the downstream SLT task. However, these approaches have a large semantic gap to solve between video features and text tokens. To address this, SignLLM~\cite{gong2024llms} proposes a tokenization strategy, converting sign videos into a language-like format such as discrete character-level sign tokens and word-level sign representations. Then, they are passed to a large language model, LLaVA \cite{liu2023visual}.

In parallel, some works have focused on visual-language pretraining methods. GFSLT-VLP \cite{zhou2023gloss} stands out as one of the first methods to leverage contrastive visual-language pretraining for gloss-free sign language translation, aligning visual features directly with textual representations. Building on this, SignCL \cite{ye2024improving} identified a representation density problem, where semantically distinct but visually similar signs tend to be close in the feature space, and proposed a contrastive learning strategy on adjacent frames to improve feature separation. Sign2GPT \cite{wong2024signgpt} proposes generating pseudo-glosses from spoken language descriptions during pretraining, offering an alternative to manual gloss annotations. Differently from these contrastive alignment methods, works such as FLa-LLM \cite{chen2024factorized} and VAP \cite{jiao2025visual} incorporate a lightweight translation model in the pretraining stage. VAP employed a multi-stage pretraining, including pseudo-gloss-to-text translation, which may be considered as weakly supervised gloss-free method.

Our approach draws inspiration from multimodal contrastive frameworks but focuses on visual–textual alignment in the context of sign language. We leverage the complementary strengths of two different visual encoders while aligning them both with each other and with language representations. This multiple-alignment strategy enables the model to capture complementary visual information from different aspects of the sign language video, leading to improved translation performance.

%% file: paper_sec/3_method.tex
\section{Method}
\begin{figure*}
  \includegraphics[width=\textwidth]{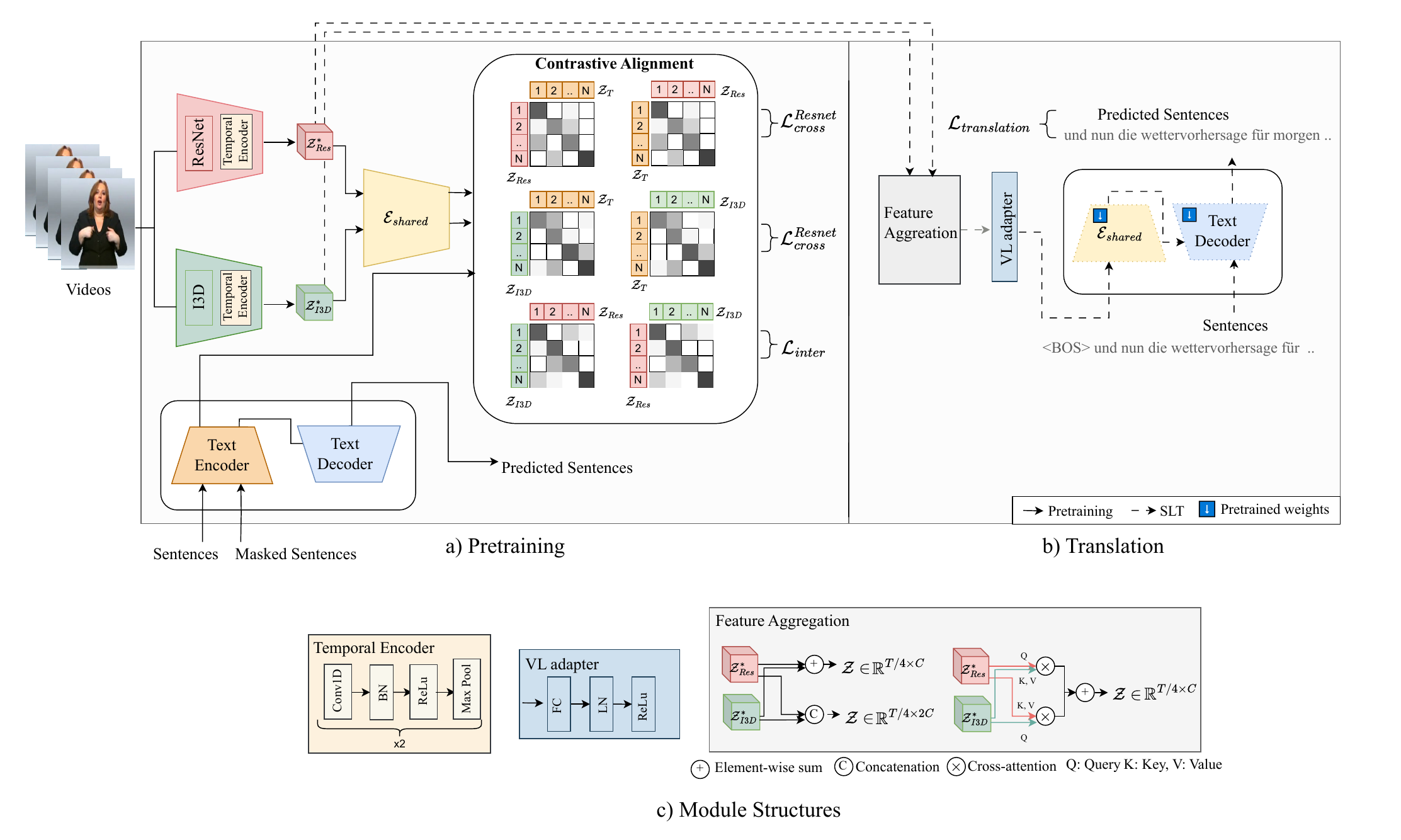}
  \caption{Overview of the proposed framework which has two phases: (a) pretraining, (b) translation. In pretraining, visual and textual representations are aligned utilizing contrastive alignment across modalities. The shared visual encoder, $\mathcal{E}_{shared}$, and Text Decoder weights are then utilized in the downstream SLT task. (c) Module structures, including the Temporal Encoder, VL Adapter, and Feature Aggregation are detailed.}
  \label{fig:network}
\end{figure*}

We propose a two-stage framework, namely DVE-SLT, for gloss-free sign language translation. Given a sign video $V=(I_1, I_2, I_3, .., I_T)$ with $T$ frames, the goal is to generate a spoken-language sentence $S=(w_1, w_2, .., w_U)$ with $U$ words without relying on gloss annotations. Our architecture is illustrated in Fig. \ref{fig:network}. In this section, we describe our pretraining strategy in which cross-modal and inter-modal alignment objectives are optimized through contrastive learning (Section \ref{sec:3.1}), and the fine-tuning stage in which the fused visual features are decoded into spoken language (Section \ref{sec:3.2}).

\subsection{Visual-Language Pretraining}
\label{sec:3.1}

    \textbf{Dual Visual Encoders.} Visual feature extraction is a critical component for video understanding tasks, including SLT. In prior works, a variety of visual encoders have been explored to extract meaningful representations. Among these, ResNet \cite{he2016deep} and I3D \cite{carreira2017quo} are frequently employed for sign language translation. ResNet-based models are commonly used for their efficiency and ability to capture fine-grained spatial features, by combining with temporal convolution layers to model temporal features \cite{chen2024factorized, ye2024improving, zhou2023gloss}. On the other hand, I3D, originally developed for action recognition, is designed to capture spatio-temporal features simultaneously. It achieved notable success in sign language recognition and translation tasks \cite{momeni2022automatic, shi2022open, sincan2023context}.

    In this work, we adopt both spatial (ResNet18) and spatio-temporal (I3D) visual encoders to extract complementary visual features. The visual encoders are followed by identical Temporal Encoders, which are composed of two blocks of a 1D-convolutional layer, a batch normalization layer, ReLu activation, and max pooling, reducing the number of frames to $T / 4$. Although I3D inherently models temporal dynamics, we apply the same temporal module to both branches to ensure consistent temporal resolution. Projecting the outputs of both branches into a unified embedding space is critical for contrastive alignment. Therefore, we employ a single shared transformer encoder, $\mathcal{E}_{shared}$, which processes both ResNet and I3D features using the same parameters. This promotes semantic consistency between modalities and enables joint optimization during pretraining.

     \textbf{Text Encoder and Decoder.} We employ the pretrained multilingual mBART \cite{liu2020-multilingual-denoising} encoder with 12 layers to convert spoken language sentences into text embeddings, which are aligned with visual features via contrastive learning. 
     
    In parallel, masked versions of the same sentences are fed into the text encoder and then passed to an mBART decoder with 3 layers, which is trained to reconstruct the original sentence using a standard cross-entropy loss. During the translation stage, this decoder is used to initialize the language decoder of our SLT model, enabling a smooth transition from pretraining to fine-tuning.

     \textbf{Contrastive Alignment Objectives.} During pretraining, we apply a \textit{cross-modal} contrastive loss to align video and text embeddings. In addition, we introduce an \textit{inter-modal} contrastive loss that aligns the representations from two complementary visual encoders for the same video. This dual-objective setup encourages both visual–text alignment and consistency across visual modalities, reinforcing the semantic coherence of the learned representations.

Given a batch of $N$ video–text pairs, we extract visual features using our shared visual encoder and textual features using a pretrained mBART multilingual text encoder. Then, for each video-text pair, we compute a similarity matrix by adopting
Cross-Lingual Contrastive Learning (CiCO) \cite{cheng2023cico}, which was proposed for the sign language retrieval task.  We use a symmetric InfoNCE loss \cite{gutmann2010noise} to contrast matching and mismatching pairs:

\begin{align}
\mathcal{L}_{\text{cross}} = & - \frac{1}{2N} \sum_{i=1}^{N} \log \frac{\exp(Sim(Z_{v_i}, Z_{t_i}) / \tau)}{\sum_{j=1}^{N} \exp(Sim(Z_{v_i}, Z_{t_j}) / \tau)} \nonumber \\
& - \frac{1}{2N} \sum_{j=1}^{N} \log \frac{\exp(Sim(Z_{v_j}, Z_{t_j}) / \tau)}{\sum_{i=1}^{N} \exp(Sim(Z_{v_i}, Z_{t_j}) / \tau)}
\end{align}

where $Sim(Z_{v_i}, Z_{t_j})$ denotes the global similarity of the $i^{th}$ video embedding and the $j^{th}$ text embedding, resulting the similarity matrix \(\mathbf{M} \in \mathbb{R}^{N \times N}\) computed over the batch, $\tau$ is a trainable temperature parameter. The visual embedding $Z_v$ is obtained from either the ResNet or I3D encoder.

In addition, since we employ two different visual encoders to extract complementary features from the same video, we introduce a second contrastive loss, \textit{inter-modal contrastive loss}. It aims to pull together the embeddings from both branches for the same input, while pushing apart mismatched visual pairs. This objective encourages semantic alignment across encoders, reinforcing the consistency of dual visual representations. Similarly, we use the symmetric InfoNCE loss \cite{gutmann2010noise} but with the visual embeddings $Z_{Res}$ and $Z_{I3D}$, extracted from the ResNet and I3D branches:

\begin{align}
\mathcal{L}_{\text{inter}} = & - \frac{1}{2N} \sum_{i=1}^{N} \log \frac{\exp(Sim(Z_{Res_i}, Z_{I3D_i}) / \tau)}{\sum_{j=1}^{N} \exp(Sim(Z_{Res_i}, Z_{I3D_j}) / \tau)} \nonumber \\
& - \frac{1}{2N} \sum_{j=1}^{N} \log \frac{\exp(Sim(Z_{I3D_j}, Z_{Res_j}) / \tau)}{\sum_{i=1}^{N} \exp(Sim(Z_{I3D_i}, Z_{Res_j}) / \tau)}
\end{align}

The total contrastive loss is the sum of two cross-modal and an inter-modal alignments:
\begin{equation}
\mathcal{L}_{\text{pretraining}} = \mathcal{L}_{\text{cross}^{Resnet}} + \mathcal{L}_{\text{cross}^{I3D}} + \mathcal{L}_{\text{inter}}
\end{equation}

where $\mathcal{L}_{\text{cross}^{Resnet}}$ and $\mathcal{L}_{\text{cross}^{I3D}}$  represent the cross-modal contrastive losses that align the visual representations extracted by the ResNet and I3D branches, respectively, with their corresponding text embeddings. The $\mathcal{L}_{\text{inter}}$  denotes an inter-modal contrastive loss that aims to pull together embeddings from both visual encoders.

\subsection{Translation}
\label{sec:3.2}
In this stage, we aim to generate accurate spoken language translations from sign language videos. We employ mBART \cite{liu2020-multilingual-denoising} as the main framework, where we initialize the encoder and decoder with the pretrained shared encoder and text decoder.

\textbf{Feature Aggregation.} After capturing visual features from both visual encoders, a critical step is the fusion of these complementary features. We explore and evaluate various feature aggregation techniques, including element-wise sum, concatenation by channel, and cross-attention as illustrated in Fig. \ref{fig:network}.

\textbf{Visual-Language Adapter.} After the aggregation of visual features, they are passed through a visual-language adapter to further refine the aggregated visual features before their integration into the mBART. It contains a fully connected layer, layer normalization, followed by a ReLu activation function. 

 We use cross-entropy loss to minimize the discrepancies between the predicted and true conditional probabilities of the spoken-language sentence given the sign language video, formally expressed as:

\begin{equation}
\mathcal{L}_{translation} = -log p(S|V)
\end{equation}

%% file: paper_sec/4_experiments.tex
\begin{table*}
    \caption{Comparison with the literature on the Phoenix2014T \cite{camgoz2018neural} test set. The best results are bold, and the second-best are underlined.}
    \centering
    \resizebox{0.9\linewidth}{!}{
    \begin{tabular}{ llcccccc }
        \hline   
         \textbf{Method} & \textbf{Publisher}  & \textbf{BLEU-1}& \textbf{BLEU-2}& \textbf{BLEU-3} & \textbf{BLEU-4} &  \textbf{ROUGE-L}   & \textbf{BLEURT} \\ \hline

        \rowcolor[HTML]{F2F2F2}
        \multicolumn{8}{c}{Gloss-based} \\ \hline
        SL-Transformer \cite{camgoz2020sign} & CVPR'20  & 46.61 & 33.73 & 26.19 & 21.32 & -& - \\
        BN-TIN-Transf.+SignBT \cite{Zhou2021CVPR} & CVPR'21  & 50.80 & 37.75 & 29.72 &24.32 & 49.54 & - \\ 
        SLTUNET \cite{zhang2022sltunet} & ICLR'22 & 52.92 & 41.76 &  33.99 & 28.47 & 52.11 & - \\ 
        TwoStream-SLT \cite{chen2022two} & NeurIPS'22 & 54.90 & 42.43 & 34.46 & 28.95 & 53.48  & - \\ 
        \hline

        \rowcolor[HTML]{F2F2F2}
        \multicolumn{8}{c}{Weakly supervised gloss-free} \\ \hline    
        VAP \cite{jiao2025visual} & ECCV'24  & 53.07 & - & - & 26.16 & 51.28 & - \\
        \hline      
        \rowcolor[HTML]{F2F2F2}
        \multicolumn{8}{c}{Gloss-free} \\ \hline 
        GFSLT-VLP \cite{zhou2023gloss} & ICCV'23    & 43.71 & 33.18 & 26.11 & 21.44  & 42.49   & - \\
        SignCL \cite{ye2024improving}  & NeurIPS'24  & \textbf{49.76} & \textbf{36.85} & \textbf{29.97} & 22.74   & \textbf{49.04} & - \\
        Sign2GPT(w/PGP) \cite{wong2024signgpt} & ICLR'24 & 49.54 & 35.96 & 28.83 & 22.52  & 48.94  & - \\
        FLa-LLM \cite{chen2024factorized} & LREC-COLING'24  & 46.29 & 35.33 & 28.03 & 23.09  & 45.27 & - \\
        SignLLM \cite{gong2024llms} & CVPR'24  & 45.21 & 34.78 & 28.05 & 23.40 & 44.49 & - \\ \hline

        DVE-SLT (Ours)  & & \underline{49.29} &  \underline{36.68} &  \underline{28.96}  & \textbf{23.81} &  \underline{48.89} & 53.59  \\  
        \hline
    \end{tabular}
    }

    \label{tab:sota} 
\end{table*}

\section{Experiments}

\subsection{Dataset and Evaluation Metrics}

\textbf{Dataset.} We evaluate on the Phoenix-2014T \cite{camgoz2018neural}, which is the most widely used sign language translation dataset. It is a German sign language dataset, containing 7096, 519, and 642 samples for training, dev, and test, with a vocabulary of 2,887 words. 

\textbf{Evaluation Metrics.} We use BLEU \cite{papineni2002bleu}, ROUGE-L \cite{lin2004rouge}, and BLEURT \cite{sellam2020bleurt} metrics to evaluate the performance of our SLT approach. BLEU measures n-gram precision and we use the sacreBLEU \cite{post2018call} implementation. ROUGE-L measures the longest common subsequence between the generated text and the reference, capturing in-order matching words without requiring consecutive overlap. BLEURT \cite{sellam2020bleurt} is a trained metric that aims to correlate human-quality scoring better. We use the BLEURT-20 checkpoint \cite{pu2021learning}. Higher scores indicate better translation. 

\subsection{Implementation Details}

Our model is implemented using PyTorch \cite{paszke2019pytorch} and trained on a single NVIDIA A100 GPU. We resize input sequences into $256\times256$ which are then cropped into $224\times224$, where random cropping is used for training and central cropping for inference. Additionally, we apply data augmentation techniques during training with a probability of 50\%, including random rotations up to 30°, random resizing with a scale factor of up to 20\%, and random translations to shift images up to 10 pixels along both the x and y axes. For pretraining, we employ an SGD optimizer with 0.9 momentum. We initialized the learning rate to 0.02 with a cosine annealing scheduler with batch size 8. For the translation stage, we explore various schedulers detailed in  Section \ref{sec:ablation}.

We initialized our I3D model with weights pre-trained on the MeineDGS German sign language dataset \cite{dgscorpus_3} for isolated sign language translation \cite{sincan2024using}. For the I3D encoder, we follow the standard setting of processing 16 consecutive frames as input. To extract features over the entire video, we apply a sliding window approach with a stride of 6 frames, resulting in temporal overlaps. To match the temporal resolution with ResNet-based embeddings, we repeat the final feature vector to fill in missing frames at the end of the sequence. 

We initialized our mBART model with \textit{mbart-large-cc25}. To save GPU memory, we trim both the model and the tokenizer based on the train set of the dataset.

\subsection{Comparison with State-of-the-art Methods}

Table~\ref{tab:sota} compares our method against existing gloss-based and gloss-free sign language translation approaches on the Phoenix-2014T benchmark. As expected, gloss-based and weakly supervised methods tend to outperform gloss-free methods due to the presence of intermediate supervision or auxiliary data. 

In the gloss-free setting, SignCL \cite{ye2024improving} and Sign2GPT \cite{wong2024signgpt} aim to learn effective representations by introducing new contrastive learning or pretraining strategies that achieve better BLEU scores with lower n-gram (BLEU-1 and BLEU-2) and ROUGE scores. On the other hand, recent large language model (LLM) based systems such as FLa-LLM \cite{chen2024factorized} and SignLLM \cite{gong2024llms}, achieve higher BLEU-4 scores, which is the most commonly used translation metric. Notably, our approach outperforms FLa-LLM \cite{chen2024factorized} and SignLLM \cite{gong2024llms} across all reported metrics, despite relying on a lighter translation model (Ours: mBART \cite{liu2020-multilingual-denoising} with 3 layers vs. mBART with 12 layers or LLaVA \cite{liu2023visual}). Notably, our model achieves the highest BLEU-4 score (23.81) among all gloss-free methods, outperforming the best competitor, SignCL, by +0.84 BLEU-4. This result highlights the strength of our dual encoder in capturing complementary features, leading to more accurate and fluent sentence-level translations.

\subsection{Ablation Studies}
\label{sec:ablation}

\begin{table}
 	\caption{Performance of different visual encoders. BN: BLEU-N, R: Rouge-L, BRT: BLEURT. }
	\centering
         \resizebox{1\linewidth}{!}{
        	\begin{tabular}{ lcccccc}
        		\hline   
        		\textbf{Encoder} & \textbf{B1} & \textbf{B2}  &  \textbf{B3} &  \textbf{B4}  &  \textbf{R} & \textbf{BRT}    \\ \hline
                ResNet \cite{he2016deep}   & 46.68 & 34.39 & 26.94 & 22.11 & 47.70 & 53.15  \\ 
                I3D \cite{carreira2017quo}   & 46.90 & 34.66 & 27.41&  22.71 & 46.69 & 52.96\\  
                ResNet + I3D   & \textbf{49.29} &  \textbf{36.68} &  \textbf{28.96}  & \textbf{23.81} &  \textbf{48.89}  & \textbf{53.59} \\
            \hline       
        	\end{tabular}
         }
     \label{tab:multimodal} 
	\end{table}


    \begin{table}
     \caption{Performance of different feature fusion techniques. BN: BLEU-N, R: Rouge-L, BRT: BLEURT. }
	\centering
         \resizebox{1\linewidth}{!}{
        	\begin{tabular}{ lcccccc}
        		\hline   
        		\textbf{Method} & \textbf{B1} & \textbf{B2} & \textbf{B3} & \textbf{B4} &  \textbf{R}  & \textbf{BRT}    \\ \hline

                Element-wise sum & 48.11 & 35.84 & 28.17 & 23.18 & 48.04 & 53.20 \\ 
                Channel-wise concat & \textbf{49.29} & \textbf{36.68} & \textbf{28.96}  & \textbf{23.81} & \textbf{48.89} & \textbf{53.59} \\
                Cross-attention & 48.09 &  35.89 & 28.53 & 23.69 & 48.74 & 53.20  \\ 
            \hline       
        	\end{tabular}
         }

     \label{tab:fusion} 
	\end{table}

    \begin{table*}
    \caption{Performance of different learning rate schedulers and hyperparameters. \textit{CosAnLR, ExpLR, OneCyleLR} stands for Cosine Annealing, Exponential, and One Cycle Learning Rate schedulers. \textit{pct\_start} represents the percentage of the training cycle spent increasing the learning rate. \textit{gamma} represents the multiplicative factor of learning rate decay. }
    \centering
    \resizebox{1\linewidth}{!}{
    \begin{tabular}{ llllcccccc }
        \hline   
          \textbf{Scheduler} & \textbf{\# Epoch} & \textbf{Batch Size}& \textbf{Hyperparams}& \textbf{BLEU-1}& \textbf{BLEU-2}& \textbf{BLEU-3} & \textbf{BLEU-4}   & \textbf{ROUGE-L}  & \textbf{BLEURT} \\ \hline

         CosAnLR & 200 & 4 & lr=0.01  & 47.54 & 34.99 & 27.53 & 22.65 & 47.77 & 53.28 \\
         CosAnLR & 200 & 8 & lr=0.02 & 48.52 & 35.91 & 28.18 & 23.18 & 48.24 & 53.56 \\
         ExpLR & 200 & 8 & lr=0.02, gamma=0.96  &  47.76 & 35.09 & 27.46 & 22.61 & 47.60 & 51.21 \\
         OneCyleLR & 150 & 8 & max lr=0.02, pct\_start=0.35 & \textbf{49.29} &  \textbf{36.68} &  \textbf{28.96}  & \textbf{23.81} &  \textbf{48.89}  & \textbf{53.39}   \\
        \hline
    \end{tabular}
    }

    \label{tab:scheduler} 
\end{table*} 


Our experiments include assessing the impact of visual encoders, feature aggregation strategies, and optimization techniques on sign language translation performance.

\textbf{Visual Encoders.} To evaluate the contributions of the two visual encoders, we conducted experiments using each encoder individually. As seen in Table \ref{tab:multimodal}, the individual performances of ResNet \cite{he2016deep} and I3D \cite{carreira2017quo} were found to be comparable, with each encoder showing slight advantages on different metrics. This suggests that both visual encoders effectively capture features relevant to sign language representation. However, fusing them by concatenating in the channel domain enhanced the results. 

\textbf{Feature aggregation techniques.} We explore several feature aggregation techniques:  element-wise sum, channel-wise concatenation, and cross-attention. As shown in Table \ref{tab:fusion}, concatenation yields the best performance across all metrics. We hypothesize that this is due to its ability to preserve the full information from both modalities. In contrast, element-wise addition may introduce mixing signals or potential information loss. We also experiment with a cross-attention mechanism, which enables the model to attend the most relevant features. However, its additional complexity might hinder optimization, especially when the two visual streams are already semantically aligned via pretraining. Nevertheless, all fusion strategies outperformed using either ResNet or I3D alone, highlighting the benefit of leveraging complementary visual features for translation.

\textbf{Effect of scheduler.} We compare the performance of different learning rate schedulers, including Cosine Annealing (CosAnLR) \cite{loshchilov2017sgdr}, Exponential (ExpLR), and One Cycle (OneCycleLR) \cite{smith2019super}, as summarized in Table \ref{tab:scheduler}. The results highlight that both CosAnLR and OneCycleLR achieved competitive performance. However, OneCycleLR demonstrates a significant advantage by achieving comparable scores in only 150 epochs, while CosAnLR requires 200 epochs. This indicates that OneCycleLR can converge faster while maintaining performance. In contrast, the Exponential LR scheduler underperformed in all metrics, suggesting that a fixed decay rate may be suboptimal for our two-stage training pipeline. 
These findings highlight the importance of selecting an appropriate learning rate scheduler for performance and efficiency.

\textbf{Qualitative analysis.} To better understand how our model attends to relevant signing segments during translation, we visualize attention patterns between video frames and generated tokens. These qualitative results complement our quantitative findings and offer insights into how well the model aligns visual and linguistic representations. 

\begin{figure}  
\centering
    \includegraphics[width=0.49\textwidth]{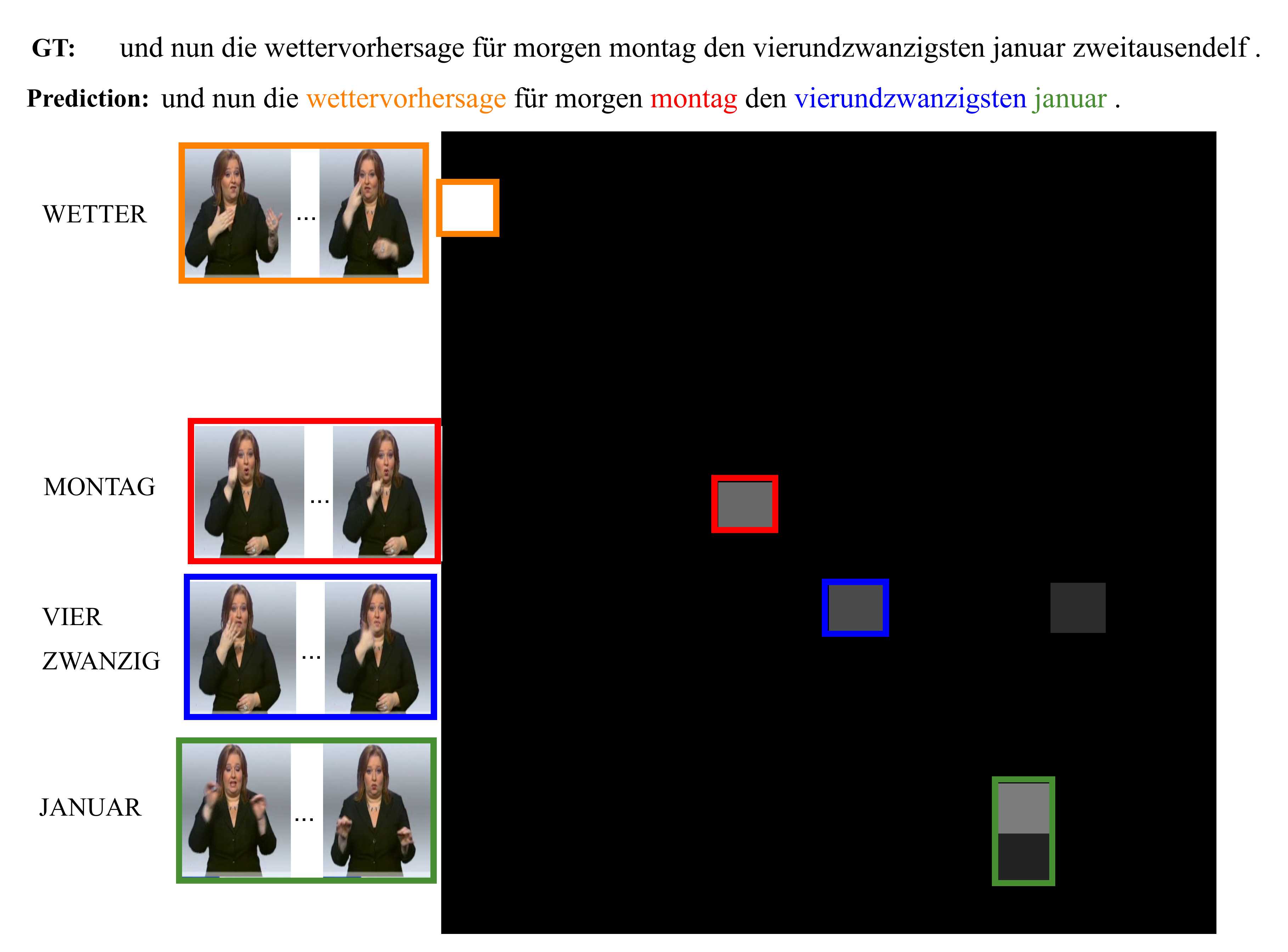}
    \caption{Cross-attention map between video frames (y-axis) and generated tokens (x-axis). Brighter regions indicate higher attention, showing that the model focuses on the correct frames. The matching colors highlight frames and tokens that align with each other.}
    \label{fig:att_map}
\end{figure}

Figure~\ref{fig:att_map} visualizes the cross-attention weights between video frames (y-axis) and generated text tokens (x-axis) during decoding. Brighter regions indicate stronger attention from the decoder to specific frames when predicting the corresponding word. To improve interpretability, we annotate the relevant video frames with the most semantically aligned tokens, using color-coded bounding boxes that match between the video and attention map.

We observe that the decoder assigns high attention scores to the correct signing segments when producing their corresponding words. For example, as shown in Figure~\ref{fig:att_map}, the token “montag” (red) aligns with the correct sign for “Monday”. The attention is temporally localized and consistent, suggesting that the model learns to ground generated text in the relevant spatio-temporal cues. These results highlight the interpretability of our architecture and its ability to align visual and linguistic semantics without gloss supervision.

%% file: paper_sec/5_conclusion.tex
\section{Conclusion}
In this paper, we propose a novel dual visual encoder framework, DVE-SLT, for sign language translation. We introduce a dual-objective contrastive alignment strategy during pretraining. This strategy encompasses cross-modal alignment to better correlate visual sign representations with textual descriptions, and inter-modal alignment that encourages consistency between the complementary features extracted by our dual visual encoders. Our approach demonstrates that combining two visual encoders improves the performance of gloss-free sign language translation by enabling the model to capture complementary visual information from different aspects of the sign language video. Future work could explore integrating additional modalities and exploring alternative pretraining strategies to further enhance translation quality.

\section*{Ethical Impact Statement}
This research aims to enhance sign language translation by leveraging deep learning techniques. While this work has the potential to improve accessibility, we are aware of potential concerns and risks. Since our model is trained on a dataset containing individuals, it is important to address privacy concerns. We used a publicly available dataset, namely Phoenix2014T \cite{camgoz2018neural}, which includes consent for non-commercial use. However, we would like to acknowledge that the dataset is limited to weather broadcast content. Although the model demonstrates high performance within this context, it might introduce biases and affect its generalizability to other domains. Additionally, we acknowledge the risk of incorrect translations, which could lead to miscommunication in practical applications. To mitigate this, we focus on obtaining more robust visual features to enhance translation accuracy.

\section*{Acknowledgements}
We would like to thank Necati Cihan Camgoz for the valuable discussions and feedback. This work was supported by the SNSF project ‘SMILE II’ (CRSII5 193686), the Innosuisse IICT Flagship (PFFS-21-47), EPSRC grant APP24554 (SignGPT-EP/Z535370/1) and and through funding from Google.org via the AI for Global Goals scheme. This work reflects only the author’s views and the funders are not responsible for any use that may be made of the information it contains.